\documentclass[letterpaper, 10 pt, conference]{ieeeconf}

\IEEEoverridecommandlockouts
\usepackage[nolist,nohyperlinks]{acronym}
\usepackage{amsmath,amsfonts,amsmath,amscd} % assumes amsmath package installed
\usepackage{amssymb}  % assumes amsmath package installed
\usepackage[backend=biber,eprint=false,mincitenames=1,maxcitenames=1,doi=false,isbn=false,url=false,style=ieee,sorting=none]{biblatex}
\usepackage{booktabs}
\usepackage{tabu}
\usepackage{multirow}
\usepackage{bm}
\usepackage[dvipsnames,table]{xcolor}
\usepackage{graphicx}
\usepackage[hidelinks,bookmarks=false]{hyperref}
\usepackage[utf8]{inputenc}
\usepackage{makecell}
\usepackage{mathtools}
\usepackage[all]{nowidow}
\usepackage{siunitx}
\usepackage{tabularx}
\usepackage{tikz}
\usepackage{tikzscale}
\usepackage[font=footnotesize]{caption}

\usepackage[normalem]{ulem}

\newcolumntype{Y}{>{\centering\arraybackslash}X}

\usepackage{mathabx}

\newcommand{\bbm}{\begin{bmatrix}}
\newcommand{\bigfh}{\widehat{\mat{F}}}
\newcommand{\ebm}{\end{bmatrix}}
\newcommand{\mat}[1]{\bm{#1}} 
\newcommand{\norm}[1]{\left\Vert#1\right\Vert}
\newcommand{\Real}{\mathbb{R}}
\newcommand{\sethree}{\mathrm{SE}(3)}
\newcommand{\algsethree}{\mathfrak{se}(3)}

\newcommand{\tinit}{\widecheck{\mat{T}}}
\newcommand{\tmean}{\bar{\mat{T}}}
\newcommand{\test}{\widehat{\mat{T}}}
\renewcommand{\vec}[1]{\bm{#1}}

\newcommand{\distrinit}{\mathcal{\bm{O}}}
\newcommand{\distrest}{\mathcal{\bm{I}}}
\newcommand{\descriptor}{\vec{d}}
\newcommand{\cov}{\mat{Y}}
\newcommand{\covest}{\widehat{\mat{Y}}}
\newcommand{\covestf}{\widehat{\mat{Y}}_{\mathrm{F}}}
\newcommand{\testf}{\widehat{\mat{T}}_{\mathrm{F}}}

\DeclareMathOperator{\tr}{tr}

\DeclareMathOperator{\adjoint}{Adj}
\DeclareMathOperator{\icp}{icp}

\title{\LARGE \bf
  CELLO-3D: Estimating the Covariance of ICP in the Real World
}

\author{David Landry \qquad François Pomerleau \qquad Philippe Giguère% <-this % stops a space
\thanks{The authors are with the Department of Computer Science and Software Engineering, Université Laval, Quebec City, Qc, G1V 0A6, Canada (emails davidlandry93@gmail.com, \{philippe.giguere,francois.pomerleau\}@ift.ulaval.ca).}}

\begin{document}

\acrodef{KL}[KL]{Kullback-Leibler}
\acrodef{CELLO}[CELLO]{Covariance Estimation and Learning through Likelihood Optimization}
\acrodef{ICP}[ICP]{Iterative Closest Point}
\acrodef{knn}[$k$-NN]{$k$-nearest neighbors}
\acrodef{SGD}[SGD]{Stochastic Gradient Descent}
\acrodef{SLAM}[SLAM]{Simultaneous Localization and Mapping}
\acrodef{DNN}[DNN]{Deep Neural Network}
\acrodefplural{DNN}[DNNs]{Deep Neural Networks}

\maketitle
\thispagestyle{empty}
\pagestyle{empty}

\begin{abstract}
The fusion of \ac{ICP} registrations in existing state estimation frameworks relies on an accurate estimation of their uncertainty.
In this paper, we study the estimation of this uncertainty in the form of a covariance.
First, we scrutinize the limitations of existing closed-form covariance estimation algorithms over 3D datasets.
Then, we set out to estimate the covariance of \ac{ICP} registrations through a data-driven approach, with over 5 100 000 registrations on 1020 pairs from real 3D point clouds.
We assess our solution upon a wide spectrum of environments, ranging from structured to unstructured and indoor to outdoor.
The capacity of our algorithm to predict covariances is accurately assessed, as well as the usefulness of these estimations for uncertainty estimation over trajectories.
The proposed method estimates covariances better than existing closed-form solutions, and makes predictions that are consistent with observed trajectories.
\end{abstract}

\section{Introduction}
The \ac{ICP} algorithm~\cite{Besl1992,Chen1991} is ubiquitous in mobile robotics for the tasks of localization and mapping.
It estimates the rigid transformation between the reference frames of two point clouds, by iteratively pairing closest points in both point clouds and minimizing a distance between those pairs.
This is equivalent to optimizing an objective function that maps rigid transformations to a scalar optimization score for a pair of point clouds. 
There is an abundance of \ac{ICP} variants~\cite{Pomerleau2015}, each of which yields slightly different transformations due to their different objective functions.
One notable variation is the choice of error metric between each pair of points, where common choices of metric are point-to-point~\cite{Besl1992} and point-to-plane~\cite{Chen1991}.
The registration process is subject to a number of sources of uncertainty and error, because of a bad adequation between the objective function and the desired result.
Chief among them is the presence of local minima in the objective function.
Other causes of uncertainty comprise noise from the range sensor, and underconstrained environments such as featureless hallways~\cite{Censi2007}.

\begin{figure}
  \includegraphics[width=0.95\columnwidth]{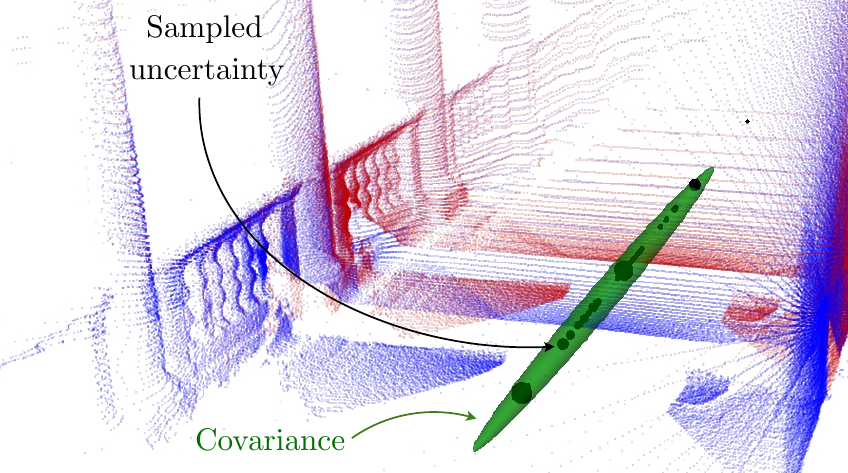}
  \caption{
    A reading point cloud (red) is being registered against a reference point cloud (blue).
    This paper studies the estimation of covariances like the green ellipsoid to allow integration of \ac{ICP} in state estimation toolchains.
    The result of \ac{ICP} was sampled, and the black balls indicate the density of registration transformations at a particular location.
    There are three larger clusters which correspond to local minimas in the objective function caused by the regularly spaced pillars.
    We project the translation part of a covariance in $\Real^3$ for illustration purposes, but in general the covariances of a 6 degrees of freedom phenomenon is studied here.
    \label{fig:problem-statement}
  }
  \vspace*{-16pt}
\end{figure}

The fusion of \ac{ICP} measurements in existing state estimation frameworks (e.g. SLAM) relies on an appropriate estimation of the uncertainty of \ac{ICP}, expressed as a covariance~\cite{Zhang1994}.
In this context, \ac{ICP} is modeled as a function of input point clouds and an initial estimate which yields a registration transformation that is normally distributed.
Optimistic covariance estimates can lead to inconsistency and navigation failures, whereas pessimistic ones inhibit efficient state estimation.
\autoref{fig:problem-statement} illustrates the process of estimating the covariance of a registration.
The registration process shown takes place in a hallway, and is consequently loosely constrained in one axis.
In this figure, optimistic covariance estimation frameworks might miss this underconstrainedness and encompass only the central samples.

This paper focuses on the problem of estimating the covariance of \ac{ICP} in such real 3D environments.
To this effect, we first provide an experimental explanation as to why current covariance estimation algorithms may perform poorly in that context.
Furthermore, we present CELLO-3D, a data-driven approach to estimating the uncertainty of 3D \ac{ICP} that works with any error metric.

\section{Related Works}

\label{sec:related-work}
There are many approaches to estimating the covariance of the \ac{ICP} algorithm, each of which must balance quality of prediction and computation time.
On one end of the spectrum, Monte-Carlo (also called brute force) algorithms such as~\cite{Bengtsson2003,Iversen2017} provide an accurate estimate \ac{ICP}'s covariance.
They consist in sampling a large number of \ac{ICP} registration transsforms, and using the covariance of the sampled results as the covariance estimation. 
If a model of the environment is available, brute-force algorithms can take sensor noise into account by simulating many scans of the environment given some noise model.
However, Monte-Carlo algorithms cannot be used online due to their high computational cost.
This limits their practicality in mobile robotics applications.

\begin{figure*}[h!]
  \includegraphics[width=\textwidth]{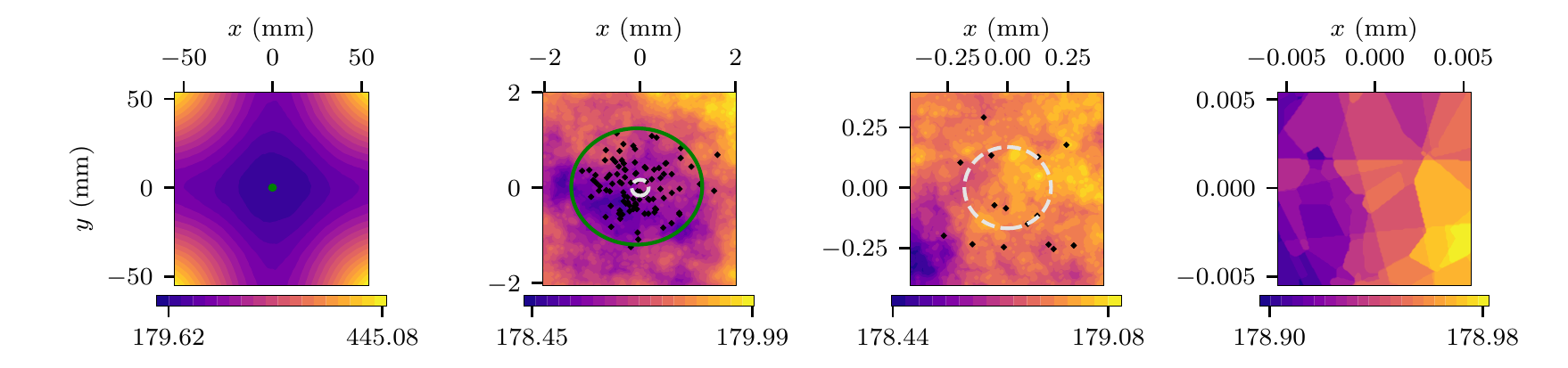}
  \caption{The objective function of point-to-plane ICP around the ground truth when registering a simple cube.
    The black dots are sampled registration transformations, with their associated translation projected on the xy plane.
    The green circle is a $3\sigma$ covariance ellipse of the distribution of registration transformations.
    The white dashed circle is a similar representation, but of Censi's covariance estimate.
    \textit{Rightmost frame}: the transitions from one plateau to another are explained by point reassociations.
    At a scale smaller than the scale of the covariance of the registration transformations, the error landscape is dominated by point reassociations.
  }
  \label{fig:error-function}
  \vspace*{-18pt}
\end{figure*}

Another important category of covariance estimation algorithms rely on the objective function's Hessian~\cite{Bengtsson2003,Biber2003,Nieto2006,Censi2007,Bosse2008,Prakhya2015}.
These closed-form methods are motivated by the need for a covariance estimation that can be used online.
Their underlying assumption is that the objective function $J(\mat{T})$ used in \ac{ICP} can be linearized around the point of convergence.
This allows the use of linear regression theory to derive a covariance from the Hessian of $J$.
If $J$ is analytically differentiable, then the Hessian can be computed directly~\cite{Bengtsson2003, Bosse2008}.
Otherwise, it can also be approximated numerically by sampling~\cite{Biber2003, Nieto2006}. %around the point of convergence
This approach accounts for errors that are due to the environment structure, but not for errors due to sensor noise.
However, modelling the effect of sensor noise on the objective function $J$ proves to be crucial for covariance estimation algorithm.
\textcite{Censi2007} addresses this by using the implicit function theorem.
His model of the covariance contains the Hessian of $J$, but also the effect of the sensor noise on it. 
It is successfully evaluated on 2D datasets.
The equations for the 3D case are derived in \cite{Prakhya2015}, but Censi's algorithm is considered to be largely optimistic in that situation~\cite{Mendes2016}.
More elaborate noise models alleviate this difficulty, but only for specific sensors~\cite{Barczyk2017}.

Closed-form approaches have the shortcoming of not taking point reassociation into account.
Therefore, they must assume that 1) \ac{ICP} converged to a loosely defined region of attraction of the ``true'' solution~\cite{Censi2007}, and 2) the reassociation of points that occur in that region have a negligible influence on the objective function.
\textcite{Bonnabel2016} show that for point-to-point \ac{ICP} variants, the second assumption is broken.
However, they present a proof that Censi's method is accurate using the point-to-plane \ac{ICP} variant in a noiseless context.
They do so by demonstrating that changes in $J$ due to point reassociations are small enough for the covariance estimation methods to remain valid under certain conditions.
In spite of this, \textcite{Mendes2016} indicates that this covariance is still optimistic in a noisy experimental context.

A data-driven alternative to covariance estimation emerged from the \ac{CELLO} framework~\cite{VegaBrown2013}.
It is a general covariance estimation strategy that projects point clouds in a descriptor space, then estimate the covariance within this space.
It uses a machine learning algorithm that first estimates a distance metric between the predictors, and then uses this metric to weigh the learning examples during online inference.
This procedure can be done with ground-truth data~\cite{VegaBrown2013}, but also without it~\cite{VegaBrown2013em} exploiting expectation-maximization.
In the presence of ground-truth data, expectation-maximization should be avoided to simplify the machine learning process.
The general strategy of \ac{CELLO} was successfully applied to the estimation of the reliability of visual features for visual-inertial navigation~\cite{Peretroukhin2015,Peretroukhin2016}.
\textcite{Peretroukhin2016} used the prediction space to generate a noise model for visual landmarks, which in turn was used to predict the ego-motion of the sensor. 
For an application to 3D \ac{ICP}, these data-driven approaches are challenging in that extracting relevant features from 3D point clouds is still an open problem.
Hand-crafted feature designers must tread carefully between the expressiveness and the generality of the descriptor for this approach to be viable.
Consequently, this method was never assessed in a 3D \ac{ICP} context to the best of our knowledge.

\vspace*{-10pt}
\section{Shortcomings of closed-form covariance estimation algorithm}
\label{ssec:shortcomings}

As discussed earlier, \textcite{Bonnabel2016} point out that closed-form covariance estimation methods are potentially ill-founded if point reassociations occur at a scale that is smaller than that of the covariance to be estimated. 
Our own analysis on 3D simulated data shows that it is likely that point reassociations happen at a scale this small.
For example, \autoref{fig:error-function} shows the objective function $J(\mat{T})$ observed when registering a pair of $1 \times 1 \times 1$ m cube shaped point clouds.
The points lie on the surface of the cube and have a $\sigma=0.01$ m noise applied on them on every axis.
At a larger scale, this objective function $J(\mat{T})$ corresponds to our intuition, with a seemingly-smooth slope towards large global minimum.
At a smaller scale, however, it is composed of a large number of ``plateaus'', each of them corresponding to one fixed association of points between the reading and the reference.
This litters the objective function with local minima to which \ac{ICP} is sensitive.
In turn, a larger covariance of \ac{ICP} results is observed.

\begin{figure}
  \centering
  \includegraphics[width=\columnwidth]{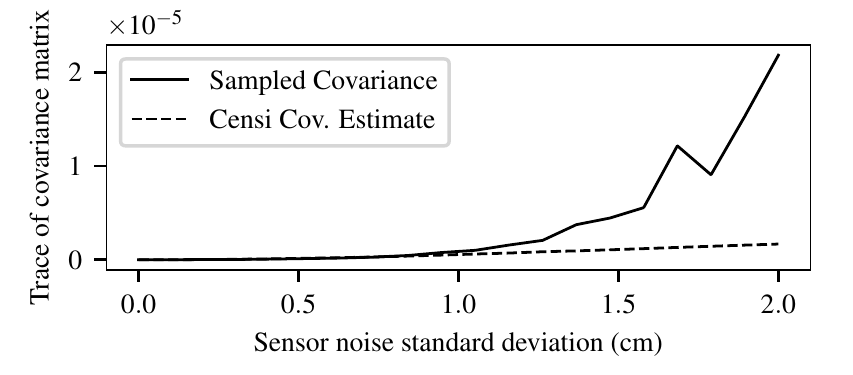}
  \caption{
    Trace of covariance computations of \ac{ICP} against sensor noise.
    Censi's covariance estimate increases slowly as the sensor noise model grows.
    The sampled covariance of \ac{ICP} increases dramatically with sensor noise, even on simulated datasets.
    This is attributed to the point reassociation provoked by sensor noise.
  \label{fig:cov-on-noise}}
\vspace{-18pt}
\end{figure}

There is a mathematical explanation for the optimism of Censi's algorithm for the point-to-point variant of \ac{ICP} in~\textcite{Bonnabel2016}.
We do not know of such a proof the point-to-plane case.
Furthermore, the results in \textcite{Bonnabel2016} were encouraging about the validity of Censi's algorithm in the point-to-plane case.
They show that the change in the objective function provoked by point reassociations is bounded under certain conditions, proving the correctness of Censi's estimate.
On the contrary, our own experiments show that caution is required even in the point-to-plane case.
Indeed, \autoref{fig:cov-on-noise} shows that sensor noise significantly impacts the empirically sampled covariance of \ac{ICP} as it grows.
On the other hand, Censi's covariance estimate grows slowly as the estimate of the sensor noise grows.
To use Censi's estimate in that experimental context, we would need to inflate our estimation of the sensor noise to values that are beyond a meaningful range.

Consequently, we argue that a viable covariance estimation algorithm for \ac{ICP} should take into account the effect of noise.
More precisely, it should model both the direct effect of the sensor noise on the objective function of \ac{ICP}, \emph{and also} the point reassociations that it provokes around ground truth.
Monte-Carlo based approaches circumvent those difficulties by incorporating the effect of noise directly.
The complexity we observe in the 3D registration process motivates our shift from analytical to data-driven solutions.
Our general approach is to implement the \ac{CELLO} framework for 3D \ac{ICP} and work towards learning covariance models from training data generated through a sampling process.
We aim at getting the best of both world: the accuracy of brute-force methods and the rapid inference of machine learning approaches.

%==============================================================
\section{3D Covariance Estimation Of ICP}
\label{sec:theory}

\begin{figure}
  \includegraphics[width=0.95\columnwidth]{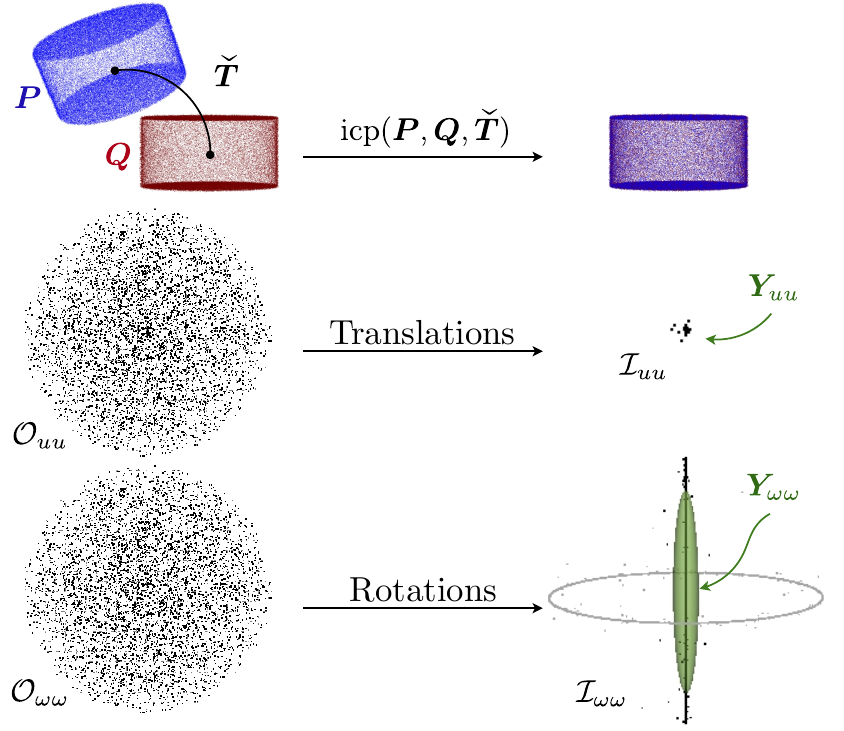}
  \caption{
  Overview of different variables used to estimate the covariance of \ac{ICP}.
  \emph{Top row:} A single example of two cylinders being registered together using \ac{ICP} and converging to well aligned point clouds $\mat{P}$ (blue) and $\mat{Q}$ (red).
  \emph{Middle row:} A view of the translation components $u$ of a set of initiation transformation $\distrinit$ before \ac{ICP} (left) and after \ac{ICP} (right).
  \emph{Bottom row:} Same view, but for the rotational components $\omega$.
  For the rotations, the spread along the vertical axis is explained by the cylinder being unconstrained around one axis.
  The presence of points in the outer ring is explained by $\mat{P}$ first turning upside down and \emph{then} being unconstrained around the same axis.
    }
    \label{fig:density-filtering}
    \vspace{-18pt}
\end{figure}

Casting \ac{CELLO} onto a 3D registration problem requires a primer on notations.
A rigid transformation $\prescript{a}{b}{\mat{T}} \in \sethree$ allows to express a point cloud $\prescript{b}{}{\mat{P}} \in \Real^{3 \times n}$ in the coordinate system $b$ in a second coordinate system $a$.
Using the Lie algebra, we can express the matrix $\mat{T}$ as a vector $\vec{\xi} \in \algsethree$ using $\log(\mat{T})$ and reverse the process using $\exp(\mat{\xi})$.
The vector $\vec{\xi}$ is split into a translation $\vec{u} \in \Real^3$ and an angle-axis rotation $\vec{\omega} \in \Real^3$.
This allows us us to express the uncertainty on a rigid transformation as a covariance matrix $\mat{Y} \in \Real^{6 \times 6}$, such that
\begin{align}
\vec{\xi} = \bbm \vec{u} \\ \vec{\omega} \ebm
&
\qquad \text{and}
&
\mat{Y} = 
\bbm 
\mat{Y}_{\vec{uu}} & \mat{Y}_{\vec{u \omega}} \\
\mat{Y}_{\vec{\omega u}} & \mat{Y}_{\vec{\omega\omega}} 
\ebm .
\end{align}
Generally speaking, we need to rely on prior information, for which we use the notation $\widecheck{(\cdot)}$, to produce an estimate $\widehat{(\cdot)}$ of a true quantity $\bar{(\cdot)}$.
For example, having access to a reference point cloud $\mat{Q} \in \Real^{3 \times m}$, it is possible to produce a transformation estimate $\test$ that reduces the alignment error between $\mat{P}$ and $\mat{Q}$ by relying on a prior transformation $\tinit$.
A typical solution to this registration problem is the \ac{ICP} algorithm:
\begin{align}
  \test = \mathrm{icp}(\mat{P}, \mat{Q}, \tinit).
\end{align}
The initial transformation $\tinit$ can be seen as an element randomly selected from a distribution of transformations $\distrinit$, for which the shape typically depend on odometry computation.
Similarly, the estimated value $\test$ comes from a distribution of transformations $\distrest$ which has a complex shape that depends on both point clouds, $\distrinit$ and the configuration of $\icp(\cdot)$.
Estimating the covariance $\mat{Y}$ of \ac{ICP} corresponds to making the (oversimplifying but tractable) assumption that the $\distrest$ is normally distributed
\begin{align}
  \distrest &\approx \mathcal{N}(\tmean, \mat{Y}),
  \label{eq:normal-approx}
\end{align}
where the right hand side is shorthand for
\begin{align}
  \tinit = \exp(\vec{\xi}) \tmean 
  &
  \qquad \text{with}
  &
  \vec{\xi} \sim \mathcal{N}(\vec{0}, \mat{Y}).
\end{align}
Note the use of $\tmean$ in \autoref{eq:normal-approx}, which supposes that \ac{ICP} is unbiased.
\autoref{fig:density-filtering} shows an example of those symbols, with two cylinders being registered starting with a wrongful initial alignment $\tinit$.
The distribution $\distrinit$ was manually configured and sampled to feed multiple $\tinit$ to \ac{ICP}.
The 5000 resulting transformations $\test$ form a distribution with covariance $\mat{Y}$ that approximates $\distrest$. 
The covariance approximates the translations well, but some points in rotations were considered divergent and filtered away.

\subsection{Covariance prediction}
\label{ssec:covariance-prediction}

\textcite{VegaBrown2013} propose a data-driven framework for covariance estimation, in which the estimation function $\bigfh$ is posed as a weighted average of training examples.
The first step in such an approach is to collect a training dataset $\mathcal{D} = \{(\descriptor_0, \mat{Y}_0), ..., (\descriptor_n, \mat{Y}_n)\}$ composed of point cloud descriptors $\descriptor_k$ and sampled covariances $\mat{Y}_k$.
Our formulation differs from the original \ac{CELLO} framework, which has error vectors in the place of the covariance matrices $\mat{Y}_k$.
This was rendered necessary due to the limited availability of 3D point cloud pairs with associated ground truth: it allows us to extract more knowledge from existing point cloud pairs.

The descriptors $\descriptor_k$ are computed by a function $\vec{g}(\prescript{a}{}{\mat{P}}, \prescript{a}{b}{\mat{T}}\prescript{b}{}{\mat{Q}})$ that extract features from a registered point cloud pair.
One can then predict a covariance $\bigfh(\descriptor)$ for an unseen example $\descriptor$ using
\begin{align}
  \bigfh(\descriptor) = \frac{1}{\sum_k s \Big( \rho(\descriptor, \descriptor_k) \Big) } \sum_{k} s \Big( \rho(\descriptor, \descriptor_k) \Big) \mat{Y}_k.
\label{eq:EstimatorF}
\end{align}
The function $\rho$ is a distance between a pair of point cloud descriptors which is defined as  
\begin{align}
  \rho(\descriptor, \descriptor') = (\descriptor - \descriptor')^\top \mat{\Theta}^\top \mat{\Theta} (\descriptor - \descriptor')
\end{align}
where $\mat{\Theta}$ is an upper triangular matrix.
The weighing function $s(x) = e^{-x}$ is chosen here, but any decreasing positive function is appropriate~\cite{VegaBrown2013}.
Large datasets could motivate a choice of $s$ that completely ignores examples with large distances to make runtime predictions more efficient~\cite{Peretroukhin2016}.
For the training of $\mat{\Theta}$, the loss for an individual covariance prediction $\bigfh(\descriptor_k)$ is
\begin{align}
  \mathcal{L}(\bigfh(\descriptor_k)|\mat{\Theta}) = \det \left( \bigfh(\descriptor_k) \right) + \tr \left( \bigfh(\descriptor_k)^{-1} \cov_k) \right) \label{eq:loss}
\end{align}
along with a regularization term~\cite{VegaBrown2013}.

\subsection{3D point cloud descriptor}
\label{ssec:description}

It is important for a point cloud descriptor $\descriptor_k$ to contain relevant information to the prediction of the covariance for $\mat{P}$, $\mat{Q}$, while being small enough to be amenable to machine learning algorithms.
Thus, the descriptor extraction function $\vec{g}(\cdot)$ should capture the geometry of the scene, in a way that translates our assumption that this geometry is an important factor of the covariance.
One consequence of this is that the extracted descriptors should \emph{not} be rotation invariant, so as to capture the correct orientation of the covariance.
For instance, the covariance of \ac{ICP} in a featureless hallway should be aligned with its walls, as depicted in \autoref{fig:problem-statement}.
There is a wide variety of descriptors that could be used for 3D point clouds~\cite{Bosse2010}, but a more thorough evaluation of the existing descriptors is a question that is left for future work.

ICP---at least in its robust version---acts mainly on the overlapping region of point clouds.
Consequently, descriptors should only be extracted from this overlapping region. In our approach, $\vec{g}$ first extracts a single point cloud $\mat{S}$ containing the subset of points from both $\mat{P}$ and $\mat{Q}$ that are overlapping, after registration.
Then, descriptors are extracted from $\mat{S}$.
The extraction pipeline separates the space into a fixed-size grid, from which all local descriptors are extracted.
First, two descriptors capture the overall ``planarity'' $p$ and ``cylindricality'' $c$ of the entire voxel, from the average values of these two metrics defined in~\cite{Demantke2011} computed at each point within the voxel.
Then, the orientation of the estimated surface normals vectors for every point in a voxel are summarized in a 9-histogram $\{h_1, ..., h_9\}$~\cite{Magnusson2009}.
The local descriptor of voxel $i$, $j$, $k$ is $\vec{v}_{ijk} = \{p,c,h_1, ..., h_9\}$.
The full descriptor $\descriptor$ for the overlaping point cloud $\mat{S}$ is the concatenation the local descriptors for every voxel.
This way, our descriptor preserves a certain amount of global information, namely the spatial distribution of the local features.

\subsection{Covariance sampling}
\label{ssec:reference-covariance}

We employ a brute-force approach similar to~\cite{Iversen2017} to estimate the covariance of \ac{ICP} for training.
We sample the result of \ac{ICP} for every point cloud pair in our dataset.
Every sample uses an initial estimate $\tinit$ drawn from $\distrinit$.
The distribution of results is then used to compute a sampled covariance $\mat{Y}_k$ for the point cloud pairs through
\begin{align}
  \cov_k = \frac{1}{n - 1}\sum_i \vec{\xi}_i \vec{\xi}_i^{\top}
\end{align}
with $\vec{\xi}_i = \log(\tmean_k^{-1}\test_i)$ the $n$ perturbations of the sampled transformations.
% Note the use of the ground truth $\tmean$ here, which enforces our supposition that \ac{ICP} is unbiased~\cite{Censi2007}.
In some cases, like in the lower row of \autoref{fig:density-filtering}, \ac{ICP} converges to many clusters.
If we suppose that we estimate the covariance of \ac{ICP} \emph{when it converges}, it is ultimately up to the designer to decide what converged from what did not.
We use the DBSCAN clustering algorithm~\cite{Schubert2017} on the set of $\vec{\xi}_i$, and keep only the points in the cluster which is closest to ground truth.
This filtering method has the benefit of avoiding unrepresentative sampled covariances $\cov$ in the training dataset.
It does so \emph{without} imposing an upper bound to the covariance of the samples.
One more look at \autoref{fig:density-filtering} illustrates the results of this procedure.
The central line of samples is explained by the fact that rotations around the $z$ axis are not constrained on this cylinder.
The outer ring corresponds to situations where \ac{ICP} converged upside-down, and then spun freely around the $z$ axis.
Our filtering strategy is able to remove results that converged incorrectly (i.e. the outer ring) while capturing the information about an underconstrained axis (i.e. the central line).

\section{Experiments}
\label{sec:experiments}

% Just because we are running out of space. I feel this short paragraph doesn't bring much 
%CELLO-3D is validated experimentally using a variety of real 3D point clouds, for environments ranging from structured to unstructured.
%The \ac{ICP} registration transformation is sampled extensively for every pair of point cloud.
%This data allows CELLO-3D's to generate the covariance estimator $\bigfh$ of \autoref{eq:EstimatorF}.

\begin{table}
  \caption{ICP pipeline used for the sampled covariance computation}
  \label{tab:icp-pipeline}
  \centering
  \begin{tabularx}{\columnwidth}{cY}% {YY}
    \toprule
    \textbf{Pipeline Element} & \textbf{Configuration} \\
    \midrule
    Point cloud filters & Maximum density, random subsampling \\
    Point matcher & $k$-d tree, 3 nearest neighbors\\
    Outlier filter & Trimmed distance (keep the closest 70\% of associations) \\
    Error minimizer & point-to-plane\\
    Transformation checkers & Max. 80 iterations\\
    \bottomrule
  \end{tabularx}
  \vspace{-18pt}
\end{table}

\subsection{Training datasets}
To be realistic, we used datasets that are representative of a wide variety of environments that a mobile robot can encounter.
Consequently, we used a subset of the \textit{Challenging data sets for point cloud registration algorithms}~\cite{Pomerleau2012Challenging}.
It comprises point clouds taken in environments ranging from structured to unstructured, and indoor to outdoor.
Every \textit{Challenging} dataset contains a sequence of $l$ point clouds $\mat{P}_i$ as well as ground truth positions $\tmean$ for them.
To generate a learning dataset $\mathcal{D} = \{(\vec{d}_0, \cov_0),...,(\vec{d}_n,\cov_n)\}$, we considered pairs of point clouds $\mat{P}_i, \mat{P}_j$ for all $i$, $j$ such that $i < j < l$ and $j - i \leq 4$.

The descriptors $\vec{d}_k$ were generated using $\vec{g}(\cdot)$.
We used a grid of $4 \times 4 \times 4$ spanning \SI{25}{\meter} in the $x$ and $y$ axes (parallel to the ground plane) and \SI{10}{\meter} in the $z$ axis (perpendicular to the ground plane)~\cite{Bosse2010}.
This grid was chosen because it encapsulates the typical spatial extent of a point cloud from the \textit{Challenging} datasets.

Each sampled covariance $\mat{Y}_k$ was computed from 5000 registrations.
Every registration had a $\distrinit$ that was centered at the ground truth and a covariance of $a\mat{I}$.
We set $a = 0.05$ to simulate a reasonable odometry scenario.
A typical \ac{ICP} registration pipeline was used, featuring the point-to-plane error metric.
\autoref{tab:icp-pipeline} describes the full registration pipeline, in terms of the framework layed down in~\cite{Pomerleau12comp}.
As discussed in \autoref{ssec:reference-covariance}, the outliers were filtered from the registration transformations to enforce the assumption that \ac{ICP} converged. 
Point cloud pairs where \ac{ICP} failed consistently were removed from the training dataset.
To do so we identified the pairs where \ac{ICP} converged on average more than \num{1}{m} or \num{1}{rad} away from ground truth.
In total, this work uses the data from about \num{5 100 000} registrations on \num{1020} pairs.
These registrations were performed on Compute Canada's computing clusters using a total of approximately 5 CPU-Years.

Data augmentation was performed to extract more knowledge from the computationally expensive sampling of the $\cov_k$.
By applying a transformation $\mat{T}$ on the point clouds before sending them in $\vec{g}(\cdot)$, we obtained new descriptors $\vec{d}_{\mathrm{aug}}$.
The sampled covariances were transformed similarily using the adjoint representation of $\mat{T}$ such that $\mat{Y}_{\mathrm{aug}} = \adjoint_{\mat{T}} \cdot \mat{Y}_k \cdot \adjoint_{\mat{T}}^\top$.
For our application we chose to perform data augmentation only by rotating the frames of reference around the $z$ axis of the reference point cloud.
The $z$ axis was chosen to preserve the 2.5D aspect of the dataset, while giving some rotation invariance to our algorithm.
Other rotations were expected to create examples that are not close to the registration pairs encountered online, such as examples where the trees of a forest are sideways.
A 6-DOF robot would warrant a more complete data augmentation.

\subsection{ICP trajectories computation}

Once the models were trained, we evaluated our covariance estimation algorithm for state estimation on indoor and outdoor trajectories.
In that spirit, we computed an \ac{ICP} odometry from our trajectory datasets.
Since the point cloud pairs are in a sequence of length $l$, we compounded their \ac{ICP} registration transformations using
%\begin{align}
 $\testf = \prescript{1}{0}{\test} \prescript{2}{1}{\test} ... \prescript{l}{l-1}{\test}$
%\end{align}
where $\testf$ is the final pose estimate of the odometry.
The distribution of initial transformations was $\mathcal{O} \approx \mathcal{N}(\prescript{i+1}{i}{\tmean},a\mat{I})$ with $a = 0.05$.
In a second step, we computed covariance estimation $\covest_k$ for each $\test_k$ using a covariance estimation model $\bigfh$.
Finally we compounded the covariances with the 4\textsuperscript{th} order approximation in~\textcite{Barfoot2014} to obtain a final covariance estimate $\covestf$.
This setup allowed us to assess the quality of the covariance estimation in context, over trajectories.

\vspace{-15pt}
\section{RESULTS}
\label{sec:results}
% This is already said in 5.A :
%The \textit{Challenging} datasets contains several trajectories, each traversing a different environment.
We trained on one trajectory dataset, while testing on one or many others that had the same type of environment/structure. 
%The groupings were done in a way to matched point clouds with the same type of environment and structure together. (this does not bring more information.)
Testing on separate (but similar) datasets was done to obtain a fair evaluation, while avoiding overly optimistic result due to overfitting.
This correspondence is detailed in \autoref{tab:losses}.
The weighting matrix $\mat{\Theta}$ was trained by stochastic gradient descent using a learning rate of \num{1e-5} using the loss from \autoref{eq:loss}.
The learning was stopped after 100 iterations or until convergence was reached.

\subsection{Single-Pair Covariance Prediction}
\label{ssec:training-losses}
First, we validated the quality of our approach, for single pairs of point clouds. 
We used as quality metric the \ac{KL} divergence between the sampled distribution and our estimated distribution (covariance).
This metric measures the amount of information lost when using the covariance $\bigfh(\vec{d}_k)$ instead of $\cov_k$ to express the distribution of ICP results $\distrest$.
Results reported in \autoref{tab:losses} are the average KL divergence over all pairs within a testing group.
For comparison, we also computed the average \ac{KL} divergence for a baseline covariance $\cov_{\mathrm{base}} = \frac{1}{n}\sum_k \mat{Y}_k$, using the $\cov_k$ of the training dataset.
We made similar computations with a Censi estimate $\cov_{\mathrm{Censi}}$, for the same point cloud pairs.
One should keep in mind that the mere prediction of the scale of the \ac{ICP} covariance has been historically challenging. 
As hinted by \autoref{fig:cov-on-noise}, Censi's covariance estimate have large divergences, since they are orders of magnitude smaller than the sampled covariances.
At worst, our approach makes predictions that have the correct order of magnitude, as seen in \autoref{tab:losses}.

Trajectory in self-similar environments will have point clouds (and covariances) that are similar to one another.
For these self-similar environment, gains over the baseline are expected to be modest.
In this situation, we observe that our predictor's parameters $\mat{\Theta}$ converges to nearly uniform weights after training.
Consequently, CELLO-3D treats training examples nearly equally, and outputs (something close to) their mean.
This phenomenon is clearly visible in \autoref{tab:losses}, where gains over the baseline are limited for \textit{Wood Autumn} and \textit{Wood Summer}.
For the \textit{Gazebo} environments, they also exhibit a certain degree of self-similarity, as measured by the low KL-divergence of the baseline.
Again, gains for our approach are modest there.
However, for the three indoor environments (apt, haupt, and stairs), we can see that they have the highest baseline KL-divergence.
This indicates that the sampled covariance varies largely throughout the trajectory, in comparison to the average (baseline) one.
As expected, it is where our algorithm makes the strongest gains.
%we can see stronger gains.
%These 3 have higher baseline KL-divergence, indicating 
%As expected, it is where our algorithm makes the strongest gains.
These gains for indoor environments are also explained by their structured nature, well-suited for our descriptors.
% Moreover, it is precisely for this kind of environment that we need to be able to have good predictions of covariance, as the others having a single constant covariance in the SLAM would not be as detrimental.

%However, for the 3 indoor environments (apt, haupt, and stairs), we can see stronger gains.
%These 3 have higher baseline KL-divergence, indicating that the sampled covariance varies largely throughout the trajectory, in comparison to the average (baseline) one.

\newcolumntype{C}[1]{>{\centering\let\newline\\\arraybackslash\hspace{0pt}}m{#1}}

\begin{table}
%\rowcolors{2}{white}{gray!10}
\taburowcolors[2]1{white..gray!10}
\setlength\tabcolsep{4.5pt}
  \caption{Loss of the \acs{CELLO} algorithm in various training scenarios.}
  \label{tab:losses}
  \centering
  \begin{tabu}{@{}XC{60pt}C{18pt}ccc@{}}
  
    \toprule
    %\textbf{Dataset} & \makecell{\textbf{Traj.}\\ \textbf{Length}\\ \textbf{(m)}}  & \makecell{\textbf{Trained}\\ \textbf{on}} & \makecell{\textbf{N.}\\\textbf{Pairs}} &  \multicolumn{3}{c}{\textbf{Avg. KL Divergence}} \\
    \multirow{ 2}{*}{\textbf{Dataset}} & 
    \multirow{ 2}{\hsize}{\centering\textbf{Trained on}} &
    \multirow{ 2}{\hsize}{\centering\textbf{N. Pairs}} &
    \multicolumn{3}{c}{\textbf{Avg. KL Divergence}} \\
    \cmidrule(l){4-6}
    &&&  Baseline  & Ours & Censi \\
    \midrule
    Apartment     & Haupt. \& Stairs	     & 1190 & 34.1          & \textbf{26.6}   & \num{9.19e7} \\
    Hauptgebaude 	& Apt \& Stairs      & 938  & 34.0          & \textbf{26.7}   & \num{2.06e8} \\
    Stairs 		    & Apt \& Haupt.      & 798  & 33.7          & \textbf{27.0}   & \num{7.65e7} \\
    Gazebo Summer & Gzb. Winter		   & 826  & 20.8	        & \textbf{19.8}   & \num{2.55e6} \\
    Gazebo Winter & Gzb. Summer 		     & 798  & 20.3	        & \textbf{18.9}	  & \num{2.25e6} \\
    Wood Autumn   & Wd Summer			   & 812  & 13.2          & \textbf{11.5}   & \num{4.94e6} \\
    Wood Summer   & Wd Autumn			   & 966	& 13.6		      & \textbf{11.3}	  & \num{3.52e7} \\
    \bottomrule
  \end{tabu}
\end{table}

\subsection{Consistency over Trajectories}

We evaluated the consistency of the estimated covariances when computing \ac{ICP} odometry trajectories in diverse locations.
This represents the fundamental situation where our predictor is used within a state-estimation algorithm. 
To visualize the error on the covariance in all dimensions at once, we resort to computing the Mahalanobis distance $D_{\text{M}}$ between $\test_{\mathrm{F}}$ and the ground truth $\prescript{l}{0}{\mat{T}}$
\begin{align}
  D_{\text{M}} = \sqrt{\vec{\xi}^\top \covestf^{-1} \vec{\xi}},
\end{align}
in which $\vec{\xi} = \log(\prescript{l}{0}{\mat{T}}^{-1}\test_{\mathrm{F}})$.
The distance $D_{\text{M}}$ can be thought of as the number of standard deviations between a sample and the mean, with a value zero meaning that $\test_{\mathrm{F}}$ was exactly on the ground truth.
\autoref{tab:consistency} shows the average $D_{\text{M}}$ of 100 trajectories for every covariance model.
It also lists the average Mahalanobis distances of the translation $\vec{u}$ against $\cov_{\vec{uu}}$ and rotation $\vec{\omega}$ against $\cov_{\vec{\omega \omega}}$ for the trajectories.
The average values of $D_{\text{M}}$ indicate that our algorithm is consistent overall.
We consider that an average $D_{\text{M}}$ above 3 indicates an optimistic covariance estimate, while an average below 1.5 insicates a pessimistic estimate.
In that sense, the covariance estimation algorithm is pessimistic for the \textit{Stairs} or \textit{Apartment} datasets.
However, not extreme values were found demonstrating a functional solution.

\begin{table}
  \caption{Final odometry error and consistency of CELLO-3D.}
  \label{tab:consistency}
  \centering
  \begin{tabularx}{\columnwidth}{Xcccccc}
    \toprule
    & \textbf{Length} & \multicolumn{2}{c}{\textbf{Translation}} & \multicolumn{2}{c}{\textbf{Rotation}} & \multirow{ 2}{*}{$D_{\text{M}}$}
    \\
    \cmidrule(l){2-2}
    \cmidrule(l){3-4}
    \cmidrule(l){5-6}
    \textbf{Traj.}  & \textbf{(m)} &  $\norm{\vec{u}}$ \textbf{(m)} & $D_{\text{M}}$ & $\norm{\vec{\omega}}$ \textbf{(rad)} & $D_{\text{M}}$ & {}
    \\
    \midrule
    Apartment    & 22 & 0.115  & 0.274  & 0.0331   & 0.160 & 0.540  \\
    Haupt. & 24 & 0.168  & 0.467  & 0.00910  & 0.346 & 1.15  \\
    Stairs       & 12 & 0.0664 & 0.0998 & 0.0127   & 0.307 & 0.592  \\
    Gzb.~Smmr  & 14 & 0.0396 & 0.278  & 0.0165   & 0.278 & 0.491  \\
    Gzb.~Wntr  & 15 & 0.0311 & 2.000  & 0.0144   & 2.90  & 2.50  \\ 
    Wd~Atmn   & 18 & 0.217  & 0.205  & 0.0178   & 0.394 & 0.405 \\
    Wd~Smmr   & 21 & 0.332  & 0.208  & 0.0299   & 0.533 & 0.762 \\
    \bottomrule
  \end{tabularx}
  \vspace{-15pt}
\end{table}

\begin{figure}
  \includegraphics[width=\columnwidth]{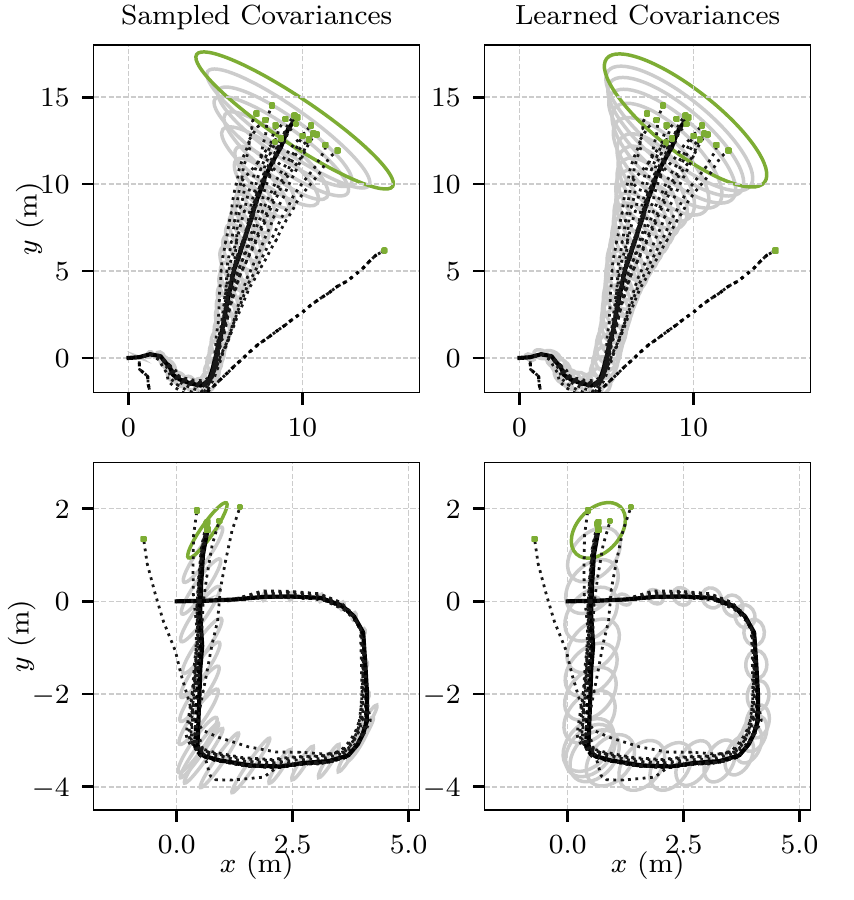}
  \caption{
    Comparison of covariance estimations for \textit{Wood Summer} (top) and \textit{Gazebo Winter} (bottom).
    Every frame as 20 \ac{ICP} odometry trajectories, although some are not visible because they are overlapping.
    The ground truth trajectory is shown as a thick black line.
    The green dots represent the final poses $\testf$, while the green ellipses represent $\covestf$.
    The data is projected on the ground plane for the sake of visualization.
    \label{fig:translation-odom}}
  \vspace{-18pt}
\end{figure}

In \autoref{fig:translation-odom}, we take a closer look at the behaviour of CELLO-3D over short trajectories, in the \textit{Wood Summer} (unstructured) and \textit{Gazebo Winter} (semi-structured) environments.
The figure compares the compounded sampled covariances $\cov_k$ with our estimated covariances $\bigfh(\vec{d}_k)$.
In there, we sampled 20 \ac{ICP} odometry trajectories to compare against the covariance predictions at each step (the grey ellipses).
The green dots represent the final $\test_{\mathrm{F}}$ for each trajectory, with the green ellipses being $\covestf$.
This figure shows that our covariance estimates are consistent with the compounded trajectories, within 2 sigma. % for the whole trajectories.
Note that the CELLO-3D uncertainty estimate grows more steadily and uniformly than that of the sampled covariances.
This indicates that, while our algorithm does not seem to model the input descriptors in very sharp detail, it is able to extract a consistent knowledge from the training dataset.

\section{Conclusion}
\label{sec:conclusion} 

In this work, we presented CELLO-3D, an online covariance estimation algorithm for \ac{ICP} that works well in 3D.
CELLO-3D uses the covariances of a learning dataset to predict the covariance of similar point cloud pairs at runtime.
It was successfully validated on individual pairs of point clouds and over trajectories, on challenging datasets.
It provides also a better uncertainty estimate when compared to existing solutions;
%Results show that covariances can be learned through CELLO-3D, providing 
Our predicted covariances are neither too optimistic nor too pessimistic, and represent well sampled particles over trajectories of several meters.

Throughout the course of this work, some challenges became visible with the transition from 2D to 3D. 
Due to the curse of dimensionality, generating a dataset for covariance in 3D requires exponentially more samples in than in 2D.
With this large number of samples, we noticed that approximating $\distrest$ as a normal distribution is prone to larger estimation errors in $\sethree$, as we observe mainly multimodal distributions 
(see \autoref{fig:problem-statement} and \autoref{fig:density-filtering}).
Moreover, the original \ac{CELLO} framework proposed the use of weak descriptors to alleviate the difficulties of descriptor design.
However, the quality of the input features is found to be critical to the success of covariance predictions.
In future works, we intend to meet this challenge using \acp{DNN} as in~\cite{Qi2017}.

%\begin{figure}
%  \includegraphics[width=\columnwidth]{img/sampling.pdf}
%  \caption{
%  Sampling results for the specific scan Apartment 03 (top) and Plain 30 (bottom) displayed in red.
%  The translation components of the registration transformation is shown as black points, with the size of the points being related to the density of samples at that location.
%  To better cover $\sethree$, 5000 initial transformations were used.
%  One can observe that the distributions are hardly normal centred on the ground truths.
%    \label{fig:problem-sampling}
%  }
%\end{figure}

\printbibliography

\end{document}